\documentclass[runningheads]{llncs}

\usepackage{eccv}

\usepackage{eccvabbrv}

\usepackage{graphicx}
\usepackage{booktabs}
\usepackage{svg}

\usepackage{booktabs}
\usepackage{xcolor}
\usepackage{colortbl}
\usepackage[normalem]{ulem}
\usepackage{multirow}

\definecolor{bestbg}{HTML}{FFCCCC}
\definecolor{secondbg}{HTML}{CCE0FF}
\newcommand{\first}[1]{\cellcolor{bestbg}\textbf{#1}}
\newcommand{\second}[1]{\cellcolor{secondbg}{#1}}

\usepackage[accsupp]{axessibility}  

\usepackage{hyperref}

\usepackage{orcidlink}

\usepackage[normalem]{ulem}   
\newif\ifshowchanges
\showchangesfalse 

\definecolor{revonecol}{RGB}{200,0,0}
\definecolor{revtwocol}{RGB}{0,90,200}
\definecolor{revthreecol}{RGB}{0,150,70}
\definecolor{newtextcol}{RGB}{0,90,200}

\newcommand{\Ra}{\ifshowchanges\textbf{\textcolor{revonecol}{[Nb8n]}}\fi}   
\newcommand{\Rb}{\ifshowchanges\textbf{\textcolor{revtwocol}{[sGgd]}}\fi}   
\newcommand{\Rc}{\ifshowchanges\textbf{\textcolor{revthreecol}{[2mJt]}}\fi} 

\newcommand{\new}[1]{\ifshowchanges\textcolor{newtextcol}{#1}\else#1\fi}

\begin{document}

\title{PanoLess: Environment Reconstruction from Partial Reflective Views} 


\author{Ahitagni Das\orcidlink{0009-0008-2981-5525} \and
Ashok Veeraraghavan\orcidlink{0000-0001-5043-7460} \and
Vivek Boominathan \orcidlink{0000-0003-4875-3135} }


\institute{Rice University, Houston TX 77005, USA}

\maketitle

\begin{abstract}
    Reflections from shiny objects and glass façades naturally extend the field of view of a camera, capturing the surrounding environment without the need to pan the camera or acquire a full panorama. We propose PanoLess, a Gaussian-splat–based framework that reconstructs the surrounding environment as a distant-illumination map from images captured on only one side of a reflective surface. PanoLess leverages surface-aligned 2D Gaussian splats with deferred shading to recover accurate per-pixel normals and reflection cues, which are fused into a neural cubemap representation of the environment. In addition, PanoLess produces a visibility map that explicitly denotes which regions of the environment are supported by the partial reflective observations. Unlike existing inverse-rendering and reflection-aware Gaussian-splatting approaches—which typically require full $360^{\circ}$ coverage and struggle under incomplete views—PanoLess enables consistent, physically grounded illumination estimation from partial-view input. We show that PanoLess achieves high-fidelity and geometrically consistent environment reconstruction, outperforming reflection-aware baselines on a new custom synthetic benchmark and publicly available datasets, and demonstrating generalization to real-world reflective captures.
\end{abstract}

\vspace{-3em}
\begin{figure}[!th]
    \centering
    \includegraphics[width=0.88\linewidth]{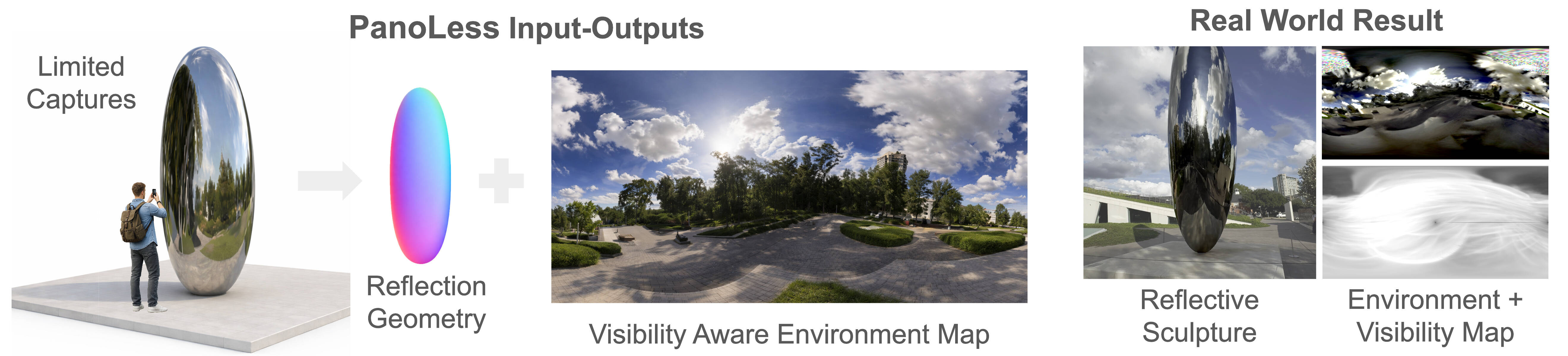}
    \caption{\textbf{PanoLess} recovers the surrounding environment from partial views of a reflective object---without panoramic capture. From images taken on one side of a shiny surface, PanoLess estimates an environment map alongside a visibility map that distinguishes well-observed regions from unsupported ones.}
    \label{fig:teaser}
\end{figure}

\vspace{-2em}
\section{Introduction}
\label{sec:intro}

Reflective surfaces such as glass façades, metals, and polished objects offer a unique opportunity to view parts of the environment that the camera itself cannot directly see. Unlike diffuse surfaces, which follow Lambertian assumptions commonly exploited in multi-view reconstruction~\cite{Ikeuchi1981,Szeliski2010}, specular reflections act as indirect imaging pathways that naturally expand the camera’s field of view without requiring panning or full-panorama capture. However, the appearance of a reflective surface is jointly determined by its geometry and the surrounding illumination, creating an inherent ambiguity: different combinations of surface normals and lighting can reproduce the same observations~\cite{Ramamoorthi2001,Basri2003}. In practical scenarios, only one side of a reflective object is typically visible, leading to partial and incomplete reflection cues that encode information about both object shape and the unseen environment. These cues are difficult to interpret reliably, and both classical multi-view stereo~\cite{Szeliski2010} and recent neural rendering pipelines~\cite{Verbin2022} struggle to disentangle geometry from illumination under such partial-view conditions.


Recovering environment maps from only partial reflective observations is a powerful means of expanding spatial awareness for robotics, AR/VR, and autonomous navigation. Yet, most existing inverse-rendering and reflection-aware methods implicitly assume dense, near-$360^{\circ}$ coverage or rely on strong supervision, making them brittle in realistic, limited-view scenarios.

Recent progress in point- and splat-based neural scene representations has opened new possibilities for real-time, high-quality rendering. Among these, Gaussian Splatting~\cite{Kerbl2023,Huang2024} has emerged as a leading paradigm for efficient scene reconstruction and differentiable rendering. However, volumetric Gaussians lack explicit surfaces and therefore produce inconsistent normals and reflection directions, limiting their ability to reason about specular phenomena. Surface-aligned 2D Gaussian splats address this by explicitly modeling tangent-plane geometry and normals, but prior extensions still rely on simplified or low-frequency lighting models that are insufficient to reconstruct the complex, high-frequency structure of real-world environment illumination—especially when large portions of the environment remain unseen.

We introduce PanoLess\footnote{Project Page: \url{https://vb-glee.github.io/panoless/}}, a framework that recovers the surrounding environment \new{as a distant-illumination map (a 2D cubemap)} from images taken on only one side of a reflective surface—without requiring large camera-panning or panoramic capture. PanoLess builds on the accurate normals and reflection cues provided by 2D Gaussian Splatting with deferred shading and aggregates them into a neural cubemap representation capable of modeling high-frequency directional radiance. Crucially, we also introduce a visibility map that quantifies which regions of the environment are directly supported by available reflections versus those that must be inferred. This explicit reasoning about missing environment regions enables consistent, physically grounded illumination estimation under highly incomplete observability. Our contributions are as follows:
\begin{itemize}
\item \new{We formulate \emph{partial-view} environment reconstruction: recovering high-fidelity environment map from reflections observed on only one side of a specular surface, without panoramic capture.}
\item \new{We introduce a \emph{visibility map}, a per-pixel confidence of the environment map, based on the rays sampled by the reflective object geometry.}
\item We introduce the Shiny Partial benchmark for partial-view environment reconstruction with ground-truth environment maps, showing PanoLess achieves nearly 5~dB improvement over baselines with sub-$5^\circ$ mean normal error.
\item We validate PanoLess on real-world handheld captures, showing it generalizes beyond synthetic data without domain-specific supervision.
\end{itemize}

\section{Related Work}
\label{sec:related-work}

\noindent\textbf{3D Gaussian Splatting and Extensions.}  
3D Gaussian Splatting (3DGS)~\cite{Kerbl2023} represents a scene with anisotropic volumetric Gaussians rasterized in real time, achieving unprecedented speed and quality for novel view synthesis. However, the original 3DGS encodes view-dependence using low-order spherical harmonics~\cite{Ramamoorthi2001}, which cannot represent sharp specular highlights. More importantly, 3D Gaussians lack explicit surfaces, leading to ambiguous surface normals and inconsistent reflections. Several extensions attempt to address these limitations. GaussianShader~\cite{Jiang2024} introduces heuristic normal estimation from Gaussian covariances to enable shading, though normals remain unstable. Relightable 3D Gaussians~\cite{Gao2024} factor appearance into diffuse and specular components with estimated normals and materials, supporting relighting. GS-IR~\cite{Liang2024} performs full inverse rendering of geometry, BRDF, and lighting within a Gaussian framework, while 3DGS-DR~\cite{Ye2024_3DGS_DR} introduces deferred reflection rendering via screen-space ray tracing. Spec-Gaussian~\cite{Yang2024} replaces spherical harmonics with anisotropic spherical Gaussians to model high-frequency lobes more faithfully. DN-Splatter~\cite{turkulainen2024dnsplatter} incorporates monocular depth and normal priors to regularize geometry. Despite these advances, all 3DGS-based approaches inherit the fundamental limitation of volumetric splats: without true surfaces, they cannot guarantee consistent normals and reflection directions, often leading to blurred or incorrect highlights.

\noindent\textbf{2D Gaussian Splatting and Reflective Surfaces.}  
2D Gaussian Splatting (2DGS)~\cite{Huang2024} overcomes the geometric ambiguity of 3DGS by introducing surface-aligned primitives: each Gaussian is a planar surfel defined by two tangent vectors, yielding explicit and consistent normals. This representation significantly improves reconstruction accuracy on standard datasets~\cite{Jensen2014}. Extensions of 2DGS include PGSR~\cite{Chen2024} (combining 2DGS with neural SDFs), GaussianSurfels~\cite{Dai2024}, Gaussian Opacity Fields~\cite{Yu2024gof}, and Street Gaussians~\cite{Yan2024}, all of which emphasize geometry fidelity. For reflective surfaces, early work such as Mirror-3DGS~\cite{Meng2024} handles planar mirrors by ray-marching, while GaussianRoom~\cite{Xiang2024} models indoor reflective scenes but relies on low-frequency lighting. Most recently, Ref-GS (CVPR 2025)~\cite{zhang2025refgs} introduces deferred shading for specular effects on 2DGS, but its lighting is implicit in the shading model. Overall, 2DGS offers a geometrically sound foundation for reflections, but current methods lack expressive environment lighting. \new{A parallel line of reflection-aware methods targets specular appearance through explicit lighting or material models—via neural or directional environment lighting~\cite{liu2023nero,ji2024reframe}, physically-based deferred rendering on 2DGS~\cite{yao2025refgaussian}, environment Gaussians and 2D Gaussian ray tracing~\cite{xie2025envgs,gu2025irgs}, or multi-view material priors~\cite{Zhang2025materialrefgs}; orthogonally, generative priors hallucinate plausible illumination from a single image~\cite{phongthawee2024diffusionlight}.}\Rc{}

\noindent\textbf{Environment Maps and Joint Illumination.}  
Environment maps are a cornerstone of rendering reflective materials~\cite{Blinn1976,Debevec1998,Debevec2000}, enabling high-quality reflections when surface normals are known. Classical models based on spherical harmonics~\cite{Sloan2002,Kautz2000} capture only low-frequency illumination, limiting realism. Neural approaches such as Neural Radiosity~\cite{Hadadan2021}, EnvMapNet~\cite{Somanath2021}, and ENVIDR~\cite{Liang2023} introduce learnable environment representations that capture high-frequency lighting. Joint optimization of geometry and illumination has been shown to improve reconstruction accuracy~\cite{Oxholm2014,Li2018,Munkberg2022}. In contrast, prior works rely on volumetric or implicit scene models, whereas our method combines surface-aligned 2DGS with neural environment maps for high-fidelity, real-time reflective reconstruction.

\section{Methodology}
\label{sec:methodology}

\begin{figure}[!th]
\vspace{-1em}
    \centering
    \includegraphics[width=\linewidth]{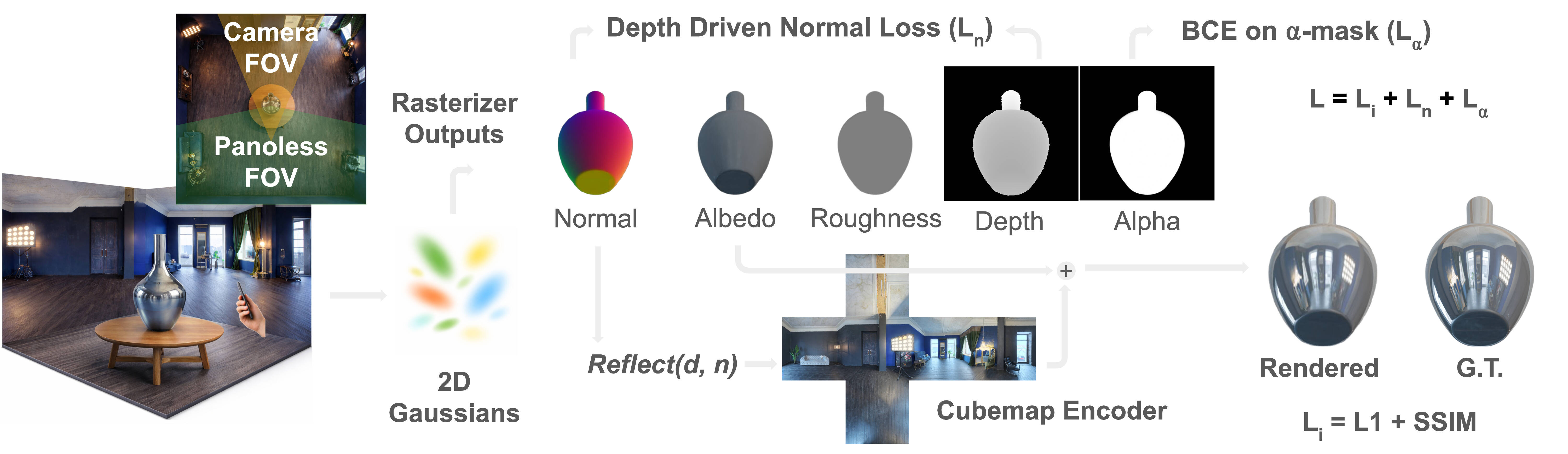}
    \caption{\textbf{Overview of our rendering pipeline.} Given a set of optimizable 2D Gaussians, we rasterize per-Gaussian properties to obtain normal, albedo, roughness, depth, and alpha maps. Surface normals and view directions give reflection directions, which query a learnable cubemap encoder to produce the specular color; this is composed with the diffuse albedo to yield the final render.}
    \label{fig:methodology}
\end{figure}

Our pipeline consists of three stages: (i)~a Gaussian splatting pass that rasterizes surface-aligned primitives into screen-space albedo and geometry buffers, (ii)~a deferred shading pass that evaluates specular reflection per pixel using a learned neural cubemap, and (iii)~a visibility accumulation pass that tracks which directions of the environment are supported by observations. An overview of our method is illustrated in Fig.~\ref{fig:methodology}.

\subsection{Scene Representation}
\label{sec:representation}

We represent the scene as a collection of $K$ surface-aligned 2D Gaussian primitives $\{G_i\}_{i=1}^{K}$, following Huang~\etal~\cite{Huang2024}. Each primitive $G_i$ is a planar surfel defined by a center $\boldsymbol{\mu}_i \in \mathbb{R}^3$, two tangent vectors $\mathbf{t}_{u,i}, \mathbf{t}_{v,i} \in \mathbb{R}^3$ that span a local tangent plane, an opacity $o_i \in [0,1]$, and a diffuse linear RGB color $\mathbf{c}_i^d \in \mathbb{R}^3$. The surface normal is given explicitly by $\mathbf{n}_i = \mathbf{t}_{u,i} \times \mathbf{t}_{v,i} / \|\mathbf{t}_{u,i} \times \mathbf{t}_{v,i}\|$. This construction guarantees that every Gaussian carries a well-defined, geometrically consistent normal, in contrast to volumetric 3D Gaussians whose shortest-axis heuristic produces unstable orientations~\cite{Kerbl2023}.

\subsection{Deferred Rasterization}
\label{sec:rasterization}

Given a camera with view matrix $\mathbf{V}$ and projection matrix $\mathbf{P}$, we project the 2D Gaussians onto the image plane and perform front-to-back alpha compositing. For a generic per-primitive attribute $\psi_i$, the composited screen-space buffer at pixel $\mathbf{p}$ is
\begin{equation}
  \Psi(\mathbf{p}) = \sum_{i \in \mathcal{N}} \psi_i \, o_i \, \mathcal{G}_i(\mathbf{p}) \prod_{j < i}\bigl(1 - o_j \, \mathcal{G}_j(\mathbf{p})\bigr),
  \label{eq:alpha-blend}
\end{equation}
where $\mathcal{N}$ denotes the depth-sorted set of primitives overlapping $\mathbf{p}$, $o_i$ is the per-primitive opacity, and $\mathcal{G}_i(\mathbf{p})$ is the screen-space Gaussian evaluation. By choosing $\psi_i$ as $\mathbf{c}_i^d$, we obtain the diffuse albedo map $\mathbf{A}$. In addition to the Albedo, the rasterizer outputs a seven-channel auxiliary map
\begin{equation}
  \mathbf{M} = \bigl[\, D_{\text{exp}},\; \alpha,\; \mathbf{N}_{\text{rast}},\; D_{\text{med}},\; D_{\text{dist}} \,\bigr] \in \mathbb{R}^{7 \times H \times W},
  \label{eq:allmap}
\end{equation}
where $D_{\text{exp}}$ and $D_{\text{med}}$ are the expected and median depth, $\alpha$ is the accumulated opacity, $\mathbf{N}_{\text{rast}} \in \mathbb{R}^3$ is the rasterized normal in camera coordinates, and $D_{\text{dist}}$ encodes depth distortion for regularization~\cite{Huang2024}.

\subsection{Surface Geometry Extraction}
\label{sec:geometry}

Robust surface geometry is critical for computing accurate reflection directions. We derive a composite surface depth by blending the expected and median depth maps:
\begin{equation}
  D_{\text{surf}}(\mathbf{p}) = (1 - \lambda_d)\, D_{\text{exp}}(\mathbf{p}) + \lambda_d\, D_{\text{med}}(\mathbf{p}),
  \label{eq:depth-blend}
\end{equation}
with $\lambda_d = 0.5$. The blending mitigates noise in expected depth caused by semi-transparent regions while retaining the sharp discontinuities captured by the median. Surface normals are then recovered by differentiating the depth map in screen space:
\begin{equation}
  \mathbf{n}_{\text{surf}}(\mathbf{p}) = \text{normalize}\!\left(\frac{\partial D_{\text{surf}}}{\partial u} \times \frac{\partial D_{\text{surf}}}{\partial v}\right),
  \label{eq:depth-normal}
\end{equation}
where $u,v$ denote image-plane coordinates. These depth-derived normals complement the rasterized normals $\mathbf{N}_{\text{rast}}$ from the alpha-composited primitive orientations; we transform the latter from camera to world space via $\mathbf{N}_{\text{world}} = \mathbf{V}_{3\times3}^{\top}\,\mathbf{N}_{\text{rast}}$ and re-normalize. In practice, we use $\mathbf{N}_{\text{world}}$ for the reflection computation and supervise $\mathbf{n}_{\text{surf}}$ as a regularizer.

\subsection{Deferred Specular Shading}
\label{sec:shading}

\new{We model the environment as distant illumination: radiance depends only on direction, so the surroundings form a function on $\mathbb{S}^2$ stored as a 2D cubemap. PanoLess thus recovers an environment map, with the Gaussian splats modeling only the reflective surface. This far-field assumption is standard in inverse rendering and accurate when the reflector is small relative to its distance from the scene, as for the fa\c{c}ades and outdoor objects we target.}\Rb{} Our shading model operates entirely in screen space, after the rasterization pass has produced the albedo and geometry buffers.

\paragraph{Reflection direction.}
Let $\mathbf{v}(\mathbf{p})$ denote the unit camera ray direction at pixel $\mathbf{p}$, obtained from the camera intrinsics and pixel coordinates. The specular reflection direction with respect to the surface normal $\mathbf{n}(\mathbf{p})$ is
\begin{equation}
  \boldsymbol{\omega}_r(\mathbf{p}) = \text{normalize}\bigl(\mathbf{v}(\mathbf{p}) - 2\,(\mathbf{v}(\mathbf{p}) \cdot \mathbf{n}(\mathbf{p}))\,\mathbf{n}(\mathbf{p})\bigr).
  \label{eq:reflection}
\end{equation}

\paragraph{Sensitivity to normal error.}
Since each cubemap query is indexed by $\boldsymbol{\omega}_r$, the fidelity of the recovered environment is governed by the accuracy of the surface normal. Perturbing the normal $\mathbf{n}\!\to\!\mathbf{n}+\delta\mathbf{n}$ at fixed view direction and dropping second-order terms, the reflection equation gives
\begin{equation}
  \delta\boldsymbol{\omega}_r = -2\bigl[(\mathbf{v}\cdot\mathbf{n})\,\delta\mathbf{n} + (\mathbf{v}\cdot\delta\mathbf{n})\,\mathbf{n}\bigr],
  \qquad
  \|\delta\boldsymbol{\omega}_r\| \le 2\bigl(|\mathbf{v}\cdot\mathbf{n}| + 1\bigr)\,\|\delta\mathbf{n}\| \le 4\,\|\delta\mathbf{n}\|.
  \label{eq:refl-sensitivity}
\end{equation}
The reflection-direction error is thus first-order linear in the normal error, with the largest amplification at grazing angles. Because pose error propagates directly into the estimated normals, this predicts a correspondingly sharp sensitivity of the recovered environment to camera-pose accuracy, which we quantify in Sec.~\ref{sec:limitations}.\Rb{}

\paragraph{Neural cubemap.}
\new{We represent the environment map as a learnable cubemap $\mathcal{E}: \mathbb{S}^2 \to \mathbb{R}^3$, stored as six trainable $L\times L$ RGB grids---one per cube face---collected in a single $6\times 3\times L\times L$ parameter tensor.}\Rb{} For a query direction $\boldsymbol{\omega} \in \mathbb{S}^2$\new{, given directly in Cartesian coordinates,} we determine the dominant face, compute the $(s,t)$ parameterization, and retrieve the radiance via bilinear interpolation \new{on the corresponding face grid}:
\begin{equation}
  \mathbf{L}_{\text{env}}(\boldsymbol{\omega}) = \sigma\!\bigl(\mathcal{E}(\boldsymbol{\omega})\bigr),
  \label{eq:envmap}
\end{equation}
where $\sigma(\cdot)$ is the sigmoid activation, constraining the output to $[0,1]$. \new{The reflection direction is fed straight to the sampler, so the specular color is exactly $\sigma(\mathcal{E}(\boldsymbol{\omega}_r))$ with no network in between.} The environment map is optimized jointly with the Gaussian parameters through the photometric loss, with gradients flowing through the bilinear sampler. Unlike methods relying on spherical harmonics~\cite{Kerbl2023} or pre-filtered mip-mapped cubemaps~\cite{Zhang2025materialrefgs}, our single-level cubemap directly stores high-frequency radiance, sufficient for mirror-like reflections. \new{Ref-GS-style encoders~\cite{Zhang2025materialrefgs} select a mip level from per-pixel roughness and decode it with an MLP; for our near-zero-roughness targets this collapses to the sharpest mip level, adding parameters without changing the output. A direct lookup gives the photometric loss an unattenuated gradient path to every texel; our ablation (Sec.~\ref{sec:ablation}) shows reintroducing roughness attenuation degrades both the render and the recovered environment---invisible under dense capture, but decisive under partial views where attenuation starves supervision exactly where samples are sparse.}

\paragraph{Color composition.}
We compose the final per-pixel color as the sum of a diffuse and a specular component:
\begin{equation}
  \mathbf{C}(\mathbf{p}) = \underbrace{\mathbf{A}(\mathbf{p})}_{\text{diffuse}} \;+\; \underbrace{\mathbf{L}_{\text{env}}\bigl(\boldsymbol{\omega}_r(\mathbf{p})\bigr)}_{\text{specular}},
  \label{eq:compose}
\end{equation}
where $\mathbf{A}$ is the rasterized albedo map. Notably, the specular term is taken directly from the environment map without an intermediate MLP or learned blending weight. This stands in contrast to prior work that modulates the environment contribution by a roughness-dependent reflection strength~\cite{Ye2024_3DGS_DR} or evaluates a full BRDF integral~\cite{Zhang2025materialrefgs}. For the highly reflective surfaces we target, a learned blending weight empirically converges to unity, so the direct connection yields cleaner gradients and faster convergence without the redundant parameterization.

\subsection{Training Objective and Protocol}
\label{sec:training}

We optimize all learnable parameters---Gaussian attributes $\{G_i\}$ and the environment map $\mathcal{E}$---end-to-end with a photometric loss. Following standard practice in Gaussian splatting~\cite{Kerbl2023,Ye2024_3DGS_DR}, we combine an $\ell_1$ term with a structural similarity term:
\begin{equation}
  \mathcal{L}_{\text{color}} = (1 - \lambda_s)\,\mathcal{L}_1 + \lambda_s\,\mathcal{L}_{\text{D-SSIM}},
  \label{eq:loss-color}
\end{equation}
where $\lambda_s = 0.2$. The individual terms are evaluated on the composited image:
\begin{align}
  \mathcal{L}_1 &= \bigl\|\, \mathbf{C}_{\text{out}} \odot \alpha + (1 - \alpha) \odot \mathbf{b} - \mathbf{I}_{\text{gt}} \,\bigr\|_1, \label{eq:l1} \\
  \mathcal{L}_{\text{D-SSIM}} &= 1 - \text{SSIM}\!\bigl(\mathbf{C}_{\text{out}} \odot \alpha + (1 - \alpha) \odot \mathbf{b},\; \mathbf{I}_{\text{gt}}\bigr), \label{eq:ssim}
\end{align}
where $\mathbf{b}$ is the background color and $\mathbf{I}_{\text{gt}}$ the ground-truth image.

We incorporate a depth-normal consistency regularizer that encourages agreement between the rasterized normals and the depth-derived normals:
\begin{equation}
  \mathcal{L}_{\text{n-d}} = \bigl\| 1 - \mathbf{N}_{\text{world}}^{\top}\, \mathbf{n}_{\text{surf}} \bigr\|_1,
  \label{eq:loss-nd}
\end{equation}
introduced by Huang~\etal~\cite{Huang2024}, which jointly sharpens the geometry and stabilizes the normal estimates that drive the reflection computation. During the first 3000 iterations, we additionally supervise the accumulated opacity with a binary cross-entropy loss against the ground-truth object mask:
\begin{equation}
  \mathcal{L}_{\alpha} = \text{BCE}\bigl(\alpha(\mathbf{p}),\; m_{\text{gt}}(\mathbf{p})\bigr),
  \label{eq:loss-alpha}
\end{equation}
where $m_{\text{gt}}$ is the ground-truth alpha mask. This early-stage supervision encourages the Gaussians to quickly converge to the object silhouette, substituting for the geometric anchoring that full-orbit multi-view consistency normally provides but which is unavailable when all cameras lie on one side. The total objective is
\begin{equation}
  \mathcal{L} = \mathcal{L}_{\text{color}} + \lambda_{\text{nd}}\,\mathcal{L}_{\text{n-d}} + [t < 3000]\,\mathcal{L}_{\alpha}.
  \label{eq:loss-total}
\end{equation}

All Gaussian parameters and the environment map $\mathcal{E}$ are optimized jointly from iteration zero. Throughout training, we apply the adaptive density control of the base 2DGS~\cite{Huang2024}, including periodic opacity reset and clone/split operations. Normal propagation~\cite{Ye2024_3DGS_DR} is employed to enlarge Gaussians, spreading well-estimated normals to neighboring primitives and accelerating convergence on reflective regions. The visibility map $\mathcal{V}$ is updated in an online fashion at each iteration with negligible overhead.

\subsection{Visibility Accumulation}
\label{sec:visibility}

A partial-view capture of a reflective object only illuminates a subset of directions in the environment. Directions that are never queried during training have no photometric supervision and must be treated differently from well-observed regions. We therefore maintain a \emph{visibility map} $\mathcal{V}: \mathbb{S}^2 \to [0,1]$ on the same cubemap grid as $\mathcal{E}$, recording the cumulative evidence for each environment direction.

During training, for every pixel $\mathbf{p}$ with $\alpha(\mathbf{p}) > \tau$, we scatter an increment to $\mathcal{V}$ at the corresponding reflection direction $\boldsymbol{\omega}_r(\mathbf{p})$, weighted by the pixel's accumulated opacity. After training, $\mathcal{V}$ provides an explicit per-texel confidence score: high values indicate directions that are well-constrained by observed reflections, while near-zero values flag regions that were never reflected into the training views. This map serves two purposes: (i)~it identifies the angular extent of the recovered environment, enabling downstream applications to distinguish trustworthy illumination from extrapolated content, and (ii)~it can be used to modulate environment map regularization, applying stronger priors in unobserved regions. This signal is meaningful only under partial coverage: with dense $360^{\circ}$
capture it saturates toward a constant, carrying information only when large portions of the environment go unsupported.

\section{Experiments}
\label{sec:experiments}

\paragraph{Datasets.}
Existing reflection benchmarks~\cite{Verbin2022} provide dense $360^{\circ}$ coverage of objects with moderate specularity and no ground-truth environment maps, precluding evaluation of partial-view environment reconstruction on highly specular surfaces. We address this gap with two benchmarks. \textbf{Shiny Partial} is a Blender-based pipeline that renders images from camera arcs confined to a single hemisphere and exports the ground-truth environment map; we release the generator alongside three fixed scenes---\textit{Cola}, \textit{Vase}, and \textit{Mirror}---with 100 training and 200 test images each. We also construct \textbf{Partial Shiny Blender} by restricting the Shiny Blender dataset~\cite{Verbin2022} to a single-hemisphere subset, providing a mixed-specularity testbed discussed in Sec.~\ref{sec:limitations}. Additionally, we validate on real-world captures (\textbf{Shiny Real}) and present ablations in Sec.~\ref{sec:ablation}.

\begin{table*}[t]
  \centering
  \caption{\textbf{Quantitative results on Shiny Partial.} We report environment map rendering quality (PSNR, SSIM, LPIPS, L1) and normal estimation accuracy (MAE, Median, RMSE in degrees, and percentage of pixels within $5^\circ$, $11.25^\circ$, and $22.5^\circ$ thresholds). PanoLess achieves the best rendering metrics on every scene and produces substantially more accurate normals, with nearly $90\%$ of pixels within $5^\circ$ of the ground truth on average. \colorbox{bestbg}{\textbf{Best}} and \colorbox{secondbg}{second best} highlighted.}
  \label{tab:shiny-partial-combined}
  \resizebox{\textwidth}{!}{%
  \begin{tabular}{ll cccc cccccc}
    \toprule
    & & \multicolumn{4}{c}{\textbf{Environment Map}} & \multicolumn{6}{c}{\textbf{Normal Estimation}} \\
    \cmidrule(lr){3-6} \cmidrule(lr){7-12}
    Scene & Method & PSNR\,$\uparrow$ & SSIM\,$\uparrow$ & LPIPS\,$\downarrow$ & L1\,$\downarrow$ & MAE\,$\downarrow$ & Median\,$\downarrow$ & RMSE\,$\downarrow$ & $<5^{\circ}$\,$\uparrow$ & $<11.25^{\circ}$\,$\uparrow$ & $<22.5^{\circ}$\,$\uparrow$ \\
    \midrule
    \multirow{5}{*}{\textit{Vase}}
      & 3DGS-DR~\cite{Ye2024_3DGS_DR}              &  7.96 & 0.204 & 0.665 & 0.346 & 14.7644          & \second{5.0592}  & 26.7193          & \second{49.82} & 63.24          & 78.73          \\
      & GShader~\cite{Jiang2024}                    &  9.62 & 0.121 & 0.668 & 0.264 & 28.4489          & 22.7388          & 34.4282          & 0.08           & 9.08           & 49.10          \\
      & Ref-Gaussian~\cite{yao2025refgaussian}           & 13.63 & 0.397 & 0.535 & \second{0.150} & 23.1096          & 17.1547          & 28.8376          & 1.03           & 27.43          & 64.48          \\
      & MaterialRefGS~\cite{Zhang2025materialrefgs} & \second{14.05} & \second{0.474} & \first{0.461} & 0.155 & \second{10.6664} & 6.2316           & \second{16.3371} & 30.39          & \second{83.10} & \second{91.87} \\
      & Ours                                        & \first{18.51} & \first{0.635} & \second{0.487} & \first{0.082} & \first{3.9683}   & \first{2.7271}   & \first{7.2938}   & \first{88.10}  & \first{96.40}  & \first{98.87}  \\
    \midrule
    \multirow{5}{*}{\textit{Cola}}
      & 3DGS-DR~\cite{Ye2024_3DGS_DR}              &  4.52 & 0.270 & 0.700 & 0.526 & 15.4632          & \second{4.4904}  & 32.4032          & \second{52.16} & 74.12          & 85.03          \\
      & GShader~\cite{Jiang2024}                    &  6.47 & 0.118 & 0.736 & 0.402 & 33.5567          & 23.7507          & 43.7941          & 0.22           & 18.70          & 48.23          \\
      & Ref-Gaussian~\cite{yao2025refgaussian}           & \second{7.26} & 0.249 & 0.660 & \second{0.372} & 22.3515          & 16.9067          & 28.5207          & 3.47           & 27.32          & 65.79          \\
      & MaterialRefGS~\cite{Zhang2025materialrefgs} &  6.81 & \second{0.302} & \second{0.587} & 0.416 & \second{10.2303} & 8.0556           & \second{13.6171} & 21.55          & \second{79.39} & \first{93.55}  \\
      & Ours                                        & \first{9.26} & \first{0.442} & \first{0.477} & \first{0.325} & \first{6.2487}   & \first{3.2135}   & \first{10.3550}  & \first{82.98}  & \first{87.48}  & \second{91.28} \\
    \midrule
    \multirow{5}{*}{\textit{Mirror}}
      & 3DGS-DR~\cite{Ye2024_3DGS_DR}              &  6.22 & 0.274 & 0.711 & 0.420 & 46.5904          & 68.1785          & 52.7519          & \second{0.11}  & 2.30           & 40.39          \\
      & GShader~\cite{Jiang2024}                    & \second{8.54} & 0.145 & \second{0.710} & \second{0.297} & 52.1474          & 52.1432          & 53.6931          & 0.00           & 0.00           & 2.25           \\
      & Ref-Gaussian~\cite{yao2025refgaussian}           &  5.56 & 0.226 & 0.746 & 0.457 & 14.7595          & 14.1869          & 15.2826          & 0.00           & 41.67          & 93.79          \\
      & MaterialRefGS~\cite{Zhang2025materialrefgs} &  5.82 & \second{0.291} & 0.755 & 0.449 & \second{14.0368} & \second{13.4597} & \second{15.0517} & 0.00           & \second{50.75} & \second{94.86} \\
      & Ours                                        & \first{13.31} & \first{0.337} & \first{0.655} & \first{0.177} & \first{4.6389}   & \first{4.6243}   & \first{4.6456}   & \first{97.00}  & \first{97.00}  & \first{97.00}  \\
    \midrule
    \multirow{5}{*}{\textit{Avg.}}
      & 3DGS-DR~\cite{Ye2024_3DGS_DR}              &  6.23 & 0.249 & 0.692 & 0.431 & 25.61            & 25.91            & 37.29            & \second{34.03} & 46.55          & 68.05          \\
      & GShader~\cite{Jiang2024}                    &  8.21 & 0.128 & 0.705 & \second{0.321} & 38.05            & 32.88            & 43.97            & 0.10           & 9.26           & 33.19          \\
      & Ref-Gaussian~\cite{yao2025refgaussian}           &  8.82 & 0.291 & 0.647 & 0.326 & 20.07            & 16.08            & 24.21            & 1.50           & 32.14          & 74.69          \\
      & MaterialRefGS~\cite{Zhang2025materialrefgs} & \second{8.89} & \second{0.356} & \second{0.601} & 0.340 & \second{11.64}   & \second{9.25}    & \second{15.00}   & 17.31          & \second{71.08} & \second{93.43} \\
      & Ours                                        & \first{13.69} & \first{0.471} & \first{0.540} & \first{0.195} & \first{4.95}     & \first{3.52}     & \first{7.43}     & \first{89.36}  & \first{93.63}  & \first{95.72}  \\
    \bottomrule
  \end{tabular}%
  }
\end{table*}

\begin{table}[t]
  \centering
  \caption{\textbf{Matched silhouette supervision.} We retrain MaterialRefGS, our strongest baseline, with our identical $\mathcal{L}_\alpha$ applied for the first 3k iterations. Even with this advantage, it trails PanoLess on every scene and metric---most starkly on \textit{Mirror}, where PanoLess places $97\%$ of normals within $5^\circ$ versus $0\%$. Normal estimation on Shiny Partial; \colorbox{bestbg}{\textbf{best}} highlighted.}
  \label{tab:matrefgs}
  \setlength{\tabcolsep}{4pt}
  \small
  \begin{tabular}{l ccc ccc}
    \toprule
    & \multicolumn{3}{c}{MaterialRefGS\,$+\,\mathcal{L}_\alpha$} & \multicolumn{3}{c}{PanoLess (Ours)} \\
    \cmidrule(lr){2-4}\cmidrule(lr){5-7}
    Scene  & MAE\,$\downarrow$ & Median\,$\downarrow$ & $<5^{\circ}$\,$\uparrow$ & MAE\,$\downarrow$ & Median\,$\downarrow$ & $<5^{\circ}$\,$\uparrow$ \\
    \midrule
    \textit{Cola}   & 10.42 &  8.43 & 20.0 & \first{6.25} & \first{3.21} & \first{83.0} \\
    \textit{Vase}   &  9.73 &  6.09 & 33.4 & \first{3.97} & \first{2.73} & \first{88.1} \\
    \textit{Mirror} & 14.79 & 13.99 &  0.0 & \first{4.64} & \first{4.62} & \first{97.0} \\
    \bottomrule
  \end{tabular}
\end{table}

\subsection{Shiny Partial Results}
\label{sec:shiny-partial-results}

\begin{figure}[!th]
    \centering
    \includegraphics[width=\linewidth]{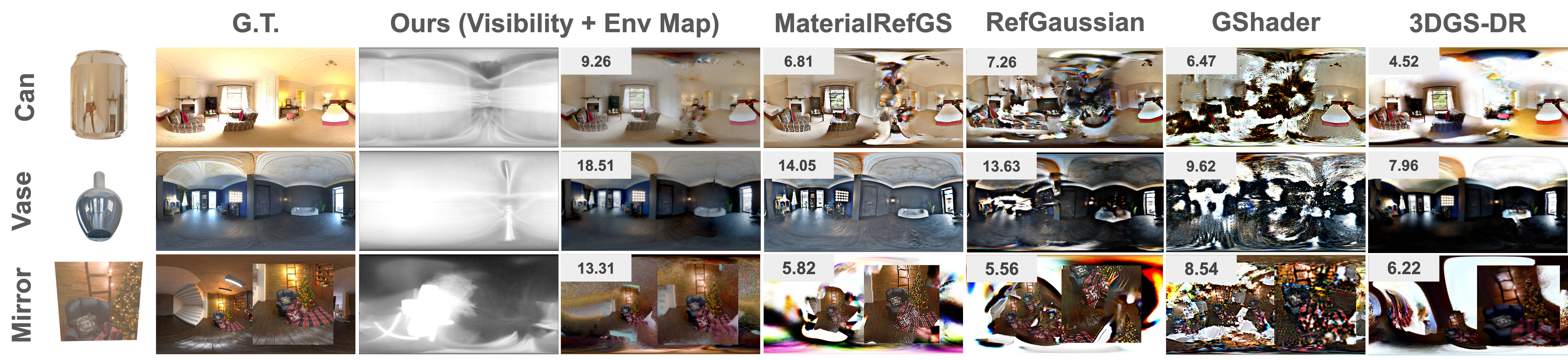}
    \caption{Recovered environment maps on \textbf{Shiny Partial}. For each scene we show the ground-truth environment map, our reconstruction, our visibility map (bright $\rightarrow$ well-observed), and the environment maps extracted from each baseline. The PSNR against the G.T. environment map is mentioned for each on the top left.}
    \label{fig:shiny-partial-env}
\end{figure}

Table~\ref{tab:shiny-partial-combined} reports quantitative results on our Shiny Partial benchmark. PanoLess outperforms all baselines on PSNR, SSIM, and L1 across every scene, and achieves the lowest LPIPS on two of three. On average, PanoLess attains 13.69~dB---a gain of nearly 5~dB over the next-best method (MaterialRefGS, 8.89~dB). The improvement is most pronounced on \textit{Mirror} (+4.8~dB), where the planar geometry requires globally consistent normals to correctly resolve the reflected environment, a setting in which volumetric baselines fail.

\begin{figure}[!th]
    \centering
    \includegraphics[width=0.73\linewidth]{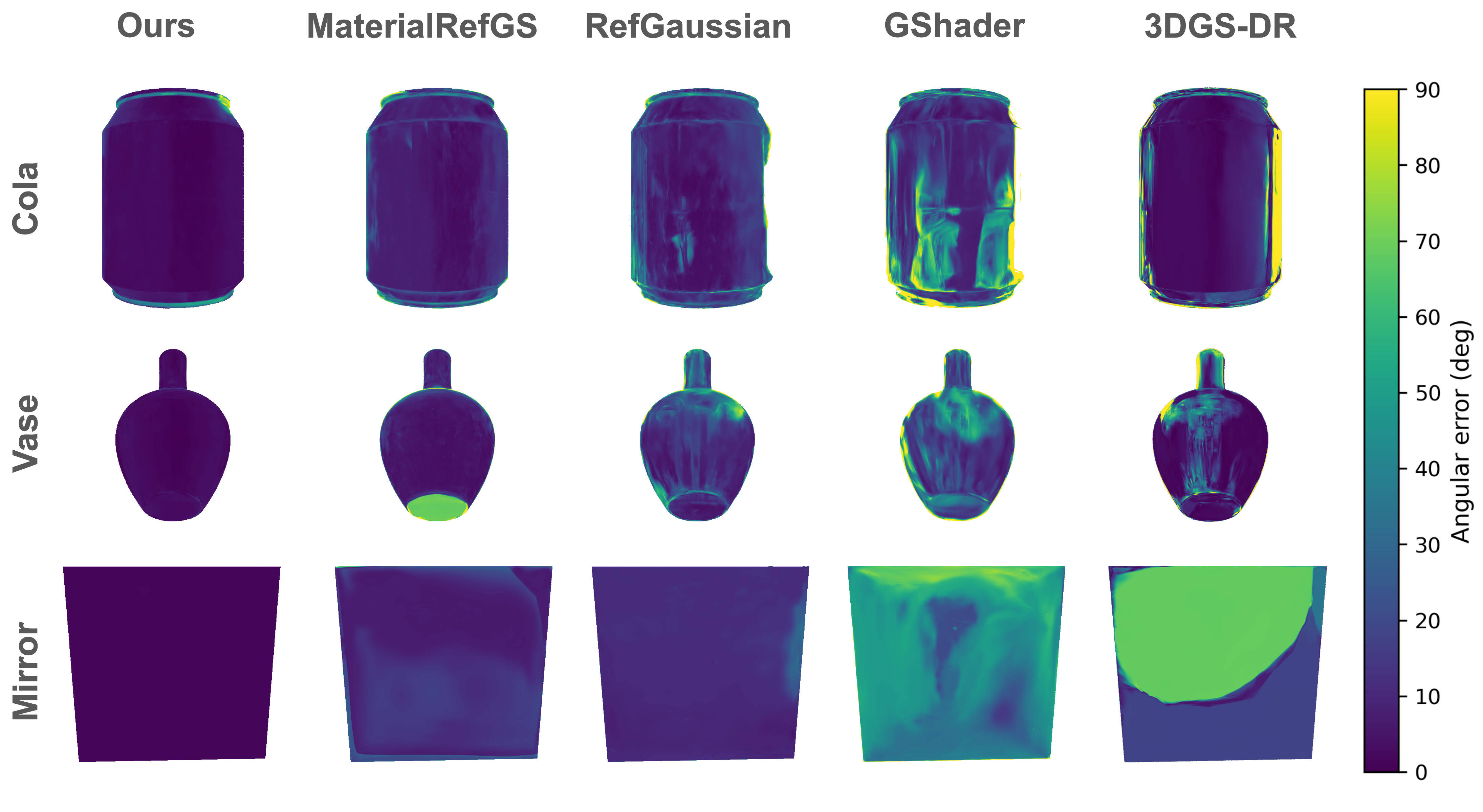}
    \caption{\textbf{Per-pixel normal angular error on Shiny Partial.} We visualize the angular error (in degrees) between predicted and ground-truth surface normals for each method across all three scenes. PanoLess produces uniformly low error across the entire surface, while baselines exhibit large errors. The color scale ranges from $0^\circ$ (dark) to $90^\circ$ (bright).}
    \label{fig:normal-error}
\end{figure}

Accurate surface normals are the single most critical factor for environment reconstruction: each pixel's reflection direction is computed directly from the estimated normal, so even small angular errors redirect the cubemap query and corrupt the recovered illumination. Table~\ref{tab:shiny-partial-combined} reports normal estimation metrics alongside rendering quality, and Fig.~\ref{fig:normal-error} visualizes per-pixel angular error across all three scenes. PanoLess achieves a mean angular error of $4.95^\circ$ on average, compared to $11.64^\circ$ for the next-best method (MaterialRefGS), and places nearly $90\%$ of pixels within $5^\circ$ of the ground truth. The improvement is especially striking on \textit{Mirror}, where PanoLess achieves $4.64^\circ$ MAE with $97\%$ of pixels below $5^\circ$, while all baselines exceed $14^\circ$ MAE. This is because planar surfaces demand globally consistent normal orientation---any local perturbation produces a visibly incorrect reflection direction. Volumetric baselines lack the surface-aligned geometry to enforce this consistency, leading to the fragmented error maps visible in Fig.~\ref{fig:normal-error}. On curved objects (\textit{Cola}, \textit{Vase}), PanoLess maintains smooth error fields that follow the surface curvature, whereas baselines show elevated error at silhouettes and high-curvature regions where the shortest-axis normal heuristic breaks down. These results validate our core design choice of building on 2D Gaussian surfels: the explicit tangent-plane normals provide the geometric foundation without which no amount of cubemap expressiveness can recover a coherent environment.

\new{\paragraph{Fairness under matched silhouette supervision.}
A natural concern is whether PanoLess benefits unfairly from its early silhouette loss $\mathcal{L}_\alpha$. To isolate this factor, we retrain MaterialRefGS---our strongest baseline---with the identical $\mathcal{L}_\alpha$ applied for the same first 3k iterations (Tab.~\ref{tab:matrefgs}). The added supervision does improve its normal consistency over its original formulation, confirming that silhouette regularization is beneficial under sparse coverage; yet MaterialRefGS still trails PanoLess on every scene and metric. The gap is widest on \textit{Mirror}, where PanoLess places $97\%$ of normals within $5^\circ$ of ground truth against $0\%$ for MaterialRefGS$+\mathcal{L}_\alpha$. The advantage of PanoLess therefore stems from its surface-aligned representation rather than from silhouette supervision alone.}\Ra{}

The fraction of the environment that can be recovered is governed by the reflector's geometry. Curved surfaces such as \textit{Cola} and \textit{Vase} act as wide-angle mirrors: a small displacement in viewpoint sweeps the reflection direction across a large angular range, so even a single-hemisphere capture illuminates a substantial portion of the surrounding environment. A planar reflector such as \textit{Mirror}, by contrast, maps each viewpoint to a narrow directional cone, and the cumulative angular coverage from the same capture configuration remains limited. The visibility maps in Fig.~\ref{fig:shiny-partial-env} corroborate this analysis: \textit{Cola} and \textit{Vase} exhibit broad high-confidence regions, whereas \textit{Mirror} shows a compact observed band with the majority of the cubemap unsupported. Despite this restricted observability, PanoLess recovers coherent illumination within the supported region, while all baselines produce incoherent environment maps irrespective of reflector curvature.

Fig.~\ref{fig:shiny-partial-env} presents the recovered environment maps. PanoLess reconstructs recognizable illumination---room structure, light sources, and color gradients are faithfully preserved---and the accompanying visibility map explicitly delineates which directions are constrained by observed reflections versus extrapolated, an output unique to our method. All baselines yield severely degraded environments, ranging from washed-out low-frequency approximations (MaterialRefGS) to entirely incoherent reconstructions (3DGS-DR, GShader). These methods assume dense, near-$360^{\circ}$ coverage with moderate specularity and lack the direct cubemap supervision pathway exploited by PanoLess.

\begin{figure}[!th]
    \centering
    \includegraphics[width=\linewidth]{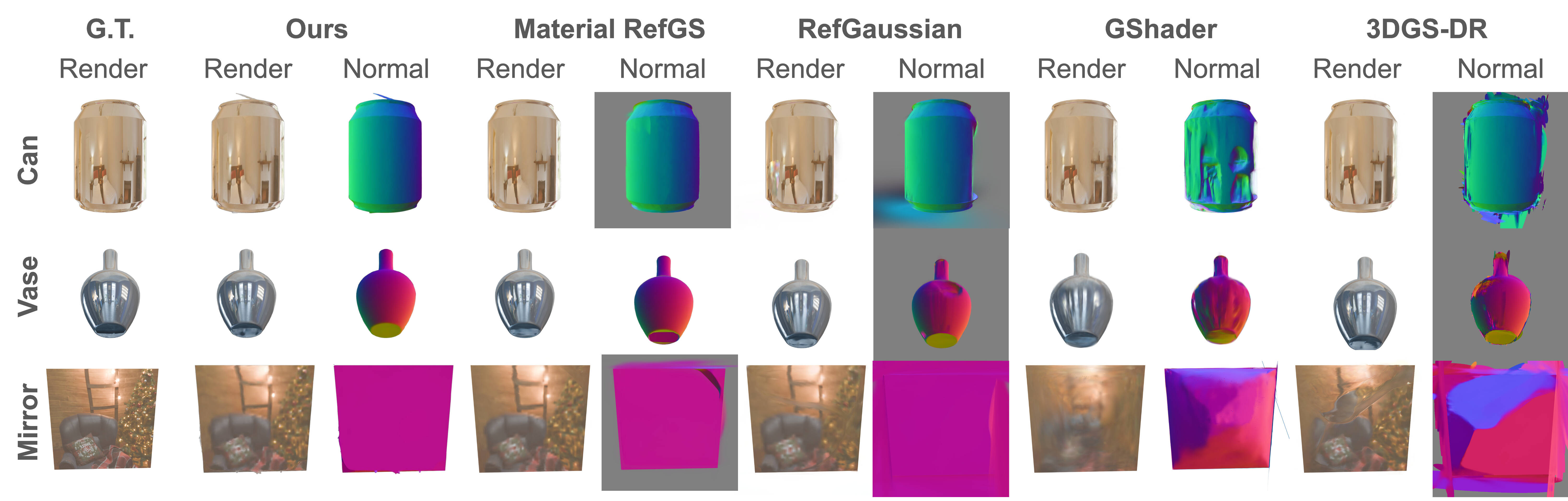}
    \caption{Novel-view renders and surface normals on \textbf{Shiny Partial}.}
    \label{fig:shiny-partial-renders}
\end{figure}

Fig.~\ref{fig:shiny-partial-renders} compares novel-view renders and surface normals. PanoLess produces smooth, geometrically accurate normal fields---clean gradients on the curved \textit{Cola} and \textit{Vase}, and a uniform orientation on the planar \textit{Mirror}---closely matching the ground truth. Baselines exhibit noisy, fragmented, or erroneous normals that propagate into reflection artifacts and degraded render quality. Ref-Gaussian and 3DGS-DR display particularly severe normal corruption on \textit{Mirror}, confirming that volumetric Gaussian representations cannot reliably model planar specular surfaces under partial-view conditions.



\subsection{Downstream Use of the Visibility Map}
\label{sec:visibility-use}
The visibility map is useful on its own: for every direction, it indicates whether the recovered illumination is backed by real observations or not. A downstream application can act on this directly, trusting the well-observed directions and treating the rest with caution without any further processing. This makes the map a natural control signal for tasks such as gating generative inpainting, applying priors only where evidence is missing, or masking unreliable directions during relighting. The Gaussian blur described next is just one simple example of using the signal, not the point of it.


As a concrete illustration, thresholding $\mathcal{V}$ (Otsu) and feathering a Gaussian blur over the flagged low-confidence directions improves \emph{every} environment-map metric on \textit{Cola} (S4). The gain is small by design---blurring only the directions the map flags as unobserved never hurts---confirming the map is well-calibrated rather than being a contribution in its own right.

\subsection{Shiny Real Results}
\label{sec:shiny-real-results}

\begin{figure}[!th]
    \centering
    \includegraphics[width=\linewidth]{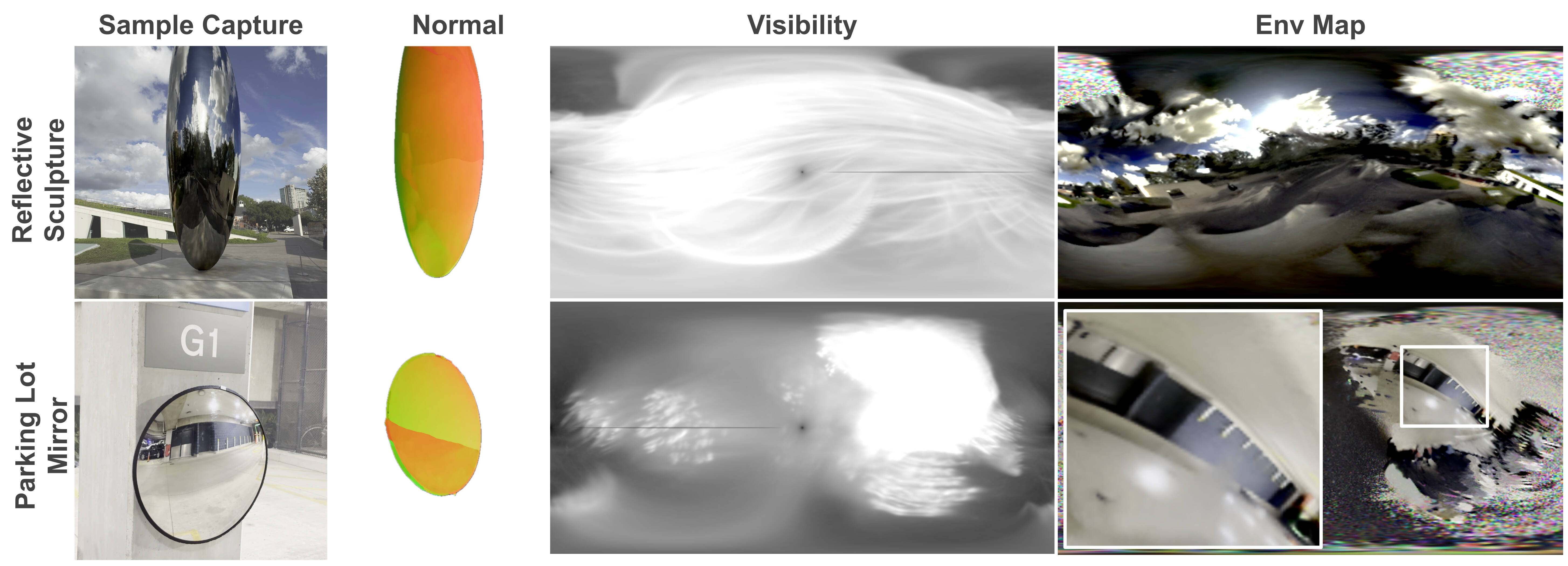}
    \caption{\textbf{Environment reconstruction on Shiny Real.} Estimated normals, visibility maps, and recovered environment maps (with PSNR/SSIM/LPIPS/L1) for a chrome column and a parking-lot mirror.}
    \label{fig:shiny-real}
\end{figure}
We capture two scenes---a reflective sculpture and a parking-lot convex mirror---with handheld video from a single side of each reflector, extracting frames and poses via COLMAP and segmenting each object with a mask. As shown in Fig.~\ref{fig:shiny-real}, PanoLess recovers recognizable illumination---sky gradients, nearby buildings, overhead lights---directly from the partial reflections without synthetic supervision, and the visibility maps correctly highlight the regions constrained by the reflector's curvature and camera trajectory.

\subsection{Ablation and Decomposition}
\label{sec:ablation}

\begin{figure}[!th]
\vspace{-3em}
    \centering
    \includegraphics[width=\linewidth]{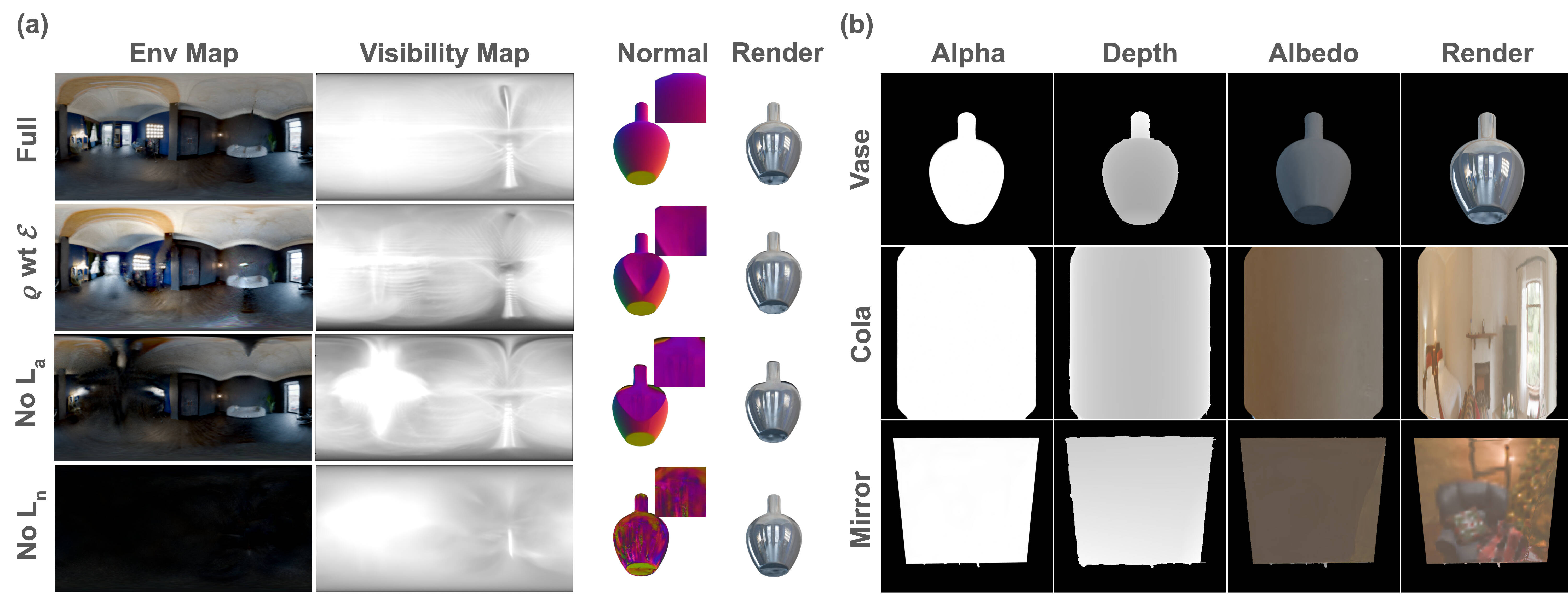}
    \caption{\textbf{(a)}~Ablation study on the \textit{Vase} scene. Each row removes or modifies one component of the full model: $\rho$-weighted~$\mathcal{E}$ scales the cubemap contribution by $(1-\rho)^2$ rather than using it directly, w/o~$\mathcal{L}_\alpha$ removes early silhouette supervision, and w/o~$\mathcal{L}_n$ removes normal consistency. We show the recovered environment map, visibility map, surface normals, and a novel-view render for each variant. \textbf{(b)}~Output decomposition of the full model on all three Shiny Partial scenes, showing the alpha mask, depth, albedo, and final render.}
    \label{fig:ablation}
\end{figure}

\begin{table}[!th]
\vspace{-1em}
  \centering
  \caption{\textbf{Ablation study} on the \textit{Vase} scene.
    We evaluate the impact of each component on novel view synthesis (Render) and environment map reconstruction (Env.\ Map).
    \colorbox{bestbg}{\textbf{Best}} and \colorbox{secondbg}{second best} highlighted.}
  \label{tab:ablation}
  \setlength{\tabcolsep}{3.5pt}
  \small
  \begin{tabular}{l cccc cccc}
    \toprule
    & \multicolumn{4}{c}{\textbf{Env.\ Map}} & \multicolumn{4}{c}{\textbf{Render}} \\
    \cmidrule(lr){2-5} \cmidrule(lr){6-9}
    Variant & PSNR\,$\uparrow$ & SSIM\,$\uparrow$ & LPIPS\,$\downarrow$ & L1\,$\downarrow$
            & PSNR\,$\uparrow$ & SSIM\,$\uparrow$ & LPIPS\,$\downarrow$ & L1\,$\downarrow$ \\
    \midrule
    Full (Ours)
      & \first{18.51} & \first{0.635} & \first{0.487} & \first{0.082}
      & \first{38.56} & \first{0.988} & \first{0.022} & \first{0.002} \\
    $\rho$-weighted $\mathcal{E}$
      & \second{13.94} & \second{0.569} & \second{0.546} & \second{0.146}
      & \second{36.16} & \second{0.982} & \second{0.036} & \second{0.003} \\
    w/o $\mathcal{L}_\alpha$
      & 12.91 & 0.466 & 0.559 & 0.177
      & 33.75 & 0.977 & 0.045 & 0.004 \\
    w/o $\mathcal{L}_n$
      & 8.68 & 0.098 & 0.621 & 0.337
      & 35.82 & 0.978 & 0.042 & 0.004 \\
    \bottomrule
  \end{tabular}
\end{table}

We ablate the key design choices of PanoLess on the \textit{Vase} scene, evaluating both novel-view synthesis and environment map reconstruction (Table~\ref{tab:ablation}, Fig.~\ref{fig:ablation}).
\paragraph{Normal consistency ($\mathcal{L}_n$).}
Removing the depth-normal consistency loss produces the most severe degradation: environment-map PSNR drops from 18.51 to 8.68~dB and SSIM collapses to 0.098. Without this regularizer, surface normals become noisy and spatially incoherent (Fig.~\ref{fig:ablation}a, bottom row), which corrupts the reflection directions queried against the cubemap. The environment map receives conflicting gradients from incorrect directions and fails to converge to a meaningful solution. This confirms that accurate geometry is the single most critical prerequisite for environment recovery---the cubemap cannot compensate for broken normals.
\paragraph{Silhouette supervision ($\mathcal{L}_\alpha$).}
Removing the early binary cross-entropy loss costs 5.6~dB in environment-map PSNR and 4.8~dB in render PSNR. Without explicit silhouette guidance during the first 3000 iterations, Gaussians are slow to consolidate at the object boundary, producing alpha leakage that contaminates normals along silhouette edges and injects erroneous reflection directions into the cubemap. The effect is visible in Fig.~\ref{fig:ablation}a as a noisier environment map and degraded normal quality near the object outline.

\paragraph{Direct cubemap supervision ($\rho$-weighted $\mathcal{E}$).}
This variant reintroduces roughness-dependent attenuation by scaling the cubemap contribution by $(1-\rho)^2$, as is standard in prior work. Since the target surfaces are highly specular, roughness values are near zero but not exactly zero, which attenuates the cubemap signal and weakens the gradient pathway from the photometric loss to the environment map. The result is a 4.6~dB drop in environment-map PSNR and visible artifacts in the recovered environment (Fig.~\ref{fig:ablation}a, second row). Setting reflection strength to unity, as our full model does, provides the cubemap with an unattenuated gradient path. The roughness parameterization is not merely redundant for highly specular objects---it is actively harmful, as even small attenuation starves the cubemap of supervision.


\paragraph{Output decomposition.}
Fig.~\ref{fig:ablation}b visualizes the intermediate outputs of the full model. The alpha masks are near-binary and depth maps smooth, confirming high-quality surfel geometry. Notably, the albedo maps are low-energy and nearly uniform---the Gaussians attribute the dominant appearance to the cubemap rather than the diffuse channel, validating the additive shading model of Eq.~\ref{eq:compose}.

\subsection{Limitations}
\label{sec:limitations}

\vspace{-1em}
\begin{figure}[!th]
    \centering
    \includegraphics[width=\linewidth]{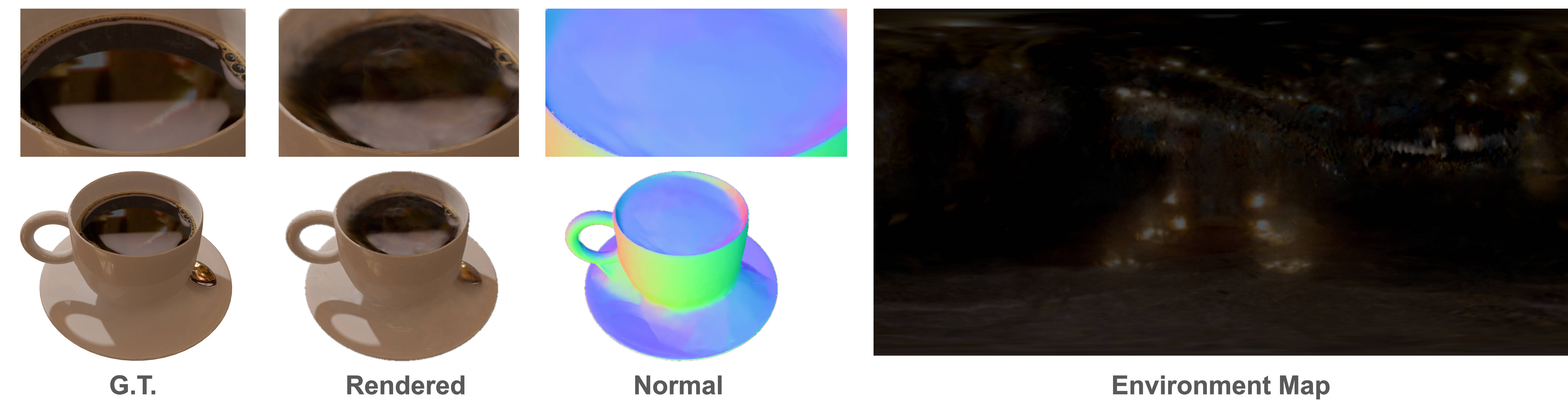}
    \caption{\textbf{Failure case.} On a coffee cup scene, the coffee surface lacks metallic reflections, providing no signal for environment reconstruction. Our method is best suited for highly specular surfaces.}
    \label{fig:limitations}
\end{figure}

PanoLess is designed for mirror-like surfaces: the unity reflection strength of Eq.~\ref{eq:compose} holds for chrome, polished metal, and glass, but breaks down as surfaces become more diffuse.

To evaluate how the method generalizes beyond this regime, we test on five scenes from the Shiny Blender dataset~\cite{Verbin2022}---\textit{Car}, \textit{Teapot}, \textit{Coffee}, \textit{Helmet}, and \textit{Toaster}---which exhibit a mixture of diffuse and specular materials. Since the original dataset provides dense $360^{\circ}$ coverage, which is not the problem we address, we construct \textbf{Partial Shiny Blender} by selecting a single-hemisphere subset of the available camera trajectories (S3), matching the partial-view conditions of our Shiny Partial benchmark.


Qualitative comparisons on Partial Shiny Blender appear in the supplementary. (S3) PanoLess continues to produce structurally consistent environment maps, albeit less coherent than on Shiny Partial: when diffuse albedo is prominent, the Gaussians absorb it into the albedo channel, diverting signal away from the cubemap and yielding a darker, less faithful reconstruction.

Fig.~\ref{fig:limitations} shows an extreme case. The matte coffee surface produces no specular reflections, so the cubemap receives zero gradient from those pixels and the recovered environment map is dim and incomplete. 
This illustrates the fundamental trade-off: the unattenuated cubemap pathway that drives PanoLess on specular surfaces fails when reflections are absent.
\textit{Sensitivity to pose error}. Because every cubemap query is indexed by a reflection direction computed from the surface normal, PanoLess is sensitive to camera-pose accuracy, as predicted by our first-order analysis (Eq.~\ref{eq:refl-sensitivity}). A pose-jitter sweep on \textit{Vase} (S5) shows envmap PSNR falling sharply from $18.4$\,dB to $12.0$\,dB at only $\sigma{=}0.5^\circ$ of camera rotation. Accurate poses---e.g.\ well-converged COLMAP reconstructions as used in our real captures---are therefore a practical prerequisite.

\section{Conclusion}
\label{sec:conclusion}

We presented PanoLess, a framework that treats highly specular surfaces as wide-angle cameras into the surrounding scene, reconstructing environment illumination from partial views without requiring camera panning or panoramic capture. By combining surface-aligned 2D Gaussian splats with a jointly optimized neural cubemap and an explicit visibility map, PanoLess recovers high-frequency environment maps from a single hemisphere of observations, achieving a nearly 5~dB improvement over the best baseline with mean normal angular error below $5^\circ$. Results on both synthetic and real-world captures confirm that the method generalizes across scenes, and the visibility map provides a principled indicator of which environment directions are trustworthy versus extrapolated, enabling informed downstream use in relighting, AR insertion, and scene understanding.

\paragraph{\textbf{Acknowledgments.}} This work was supported in part by the Ken Kennedy Institute at Rice University.

%
%
\bibliographystyle{splncs04}
\bibliography{main}

\clearpage
\appendix

\renewcommand{\thesection}{S\arabic{section}}
\renewcommand{\thetable}{S\arabic{table}}
\renewcommand{\thefigure}{S\arabic{figure}}
\renewcommand{\theequation}{S\arabic{equation}}
\setcounter{section}{0}
\setcounter{figure}{0}
\setcounter{table}{0}

\setlength{\textfloatsep}{10pt plus 2pt minus 2pt}
\setlength{\intextsep}{10pt plus 2pt minus 2pt}
\setlength{\floatsep}{10pt plus 2pt minus 2pt}

\section{Shiny Real Capture Details}
\label{sec:shiny-real-capture}

\begin{figure}[!th]
    \centering
    \includegraphics[width=0.68 \linewidth]{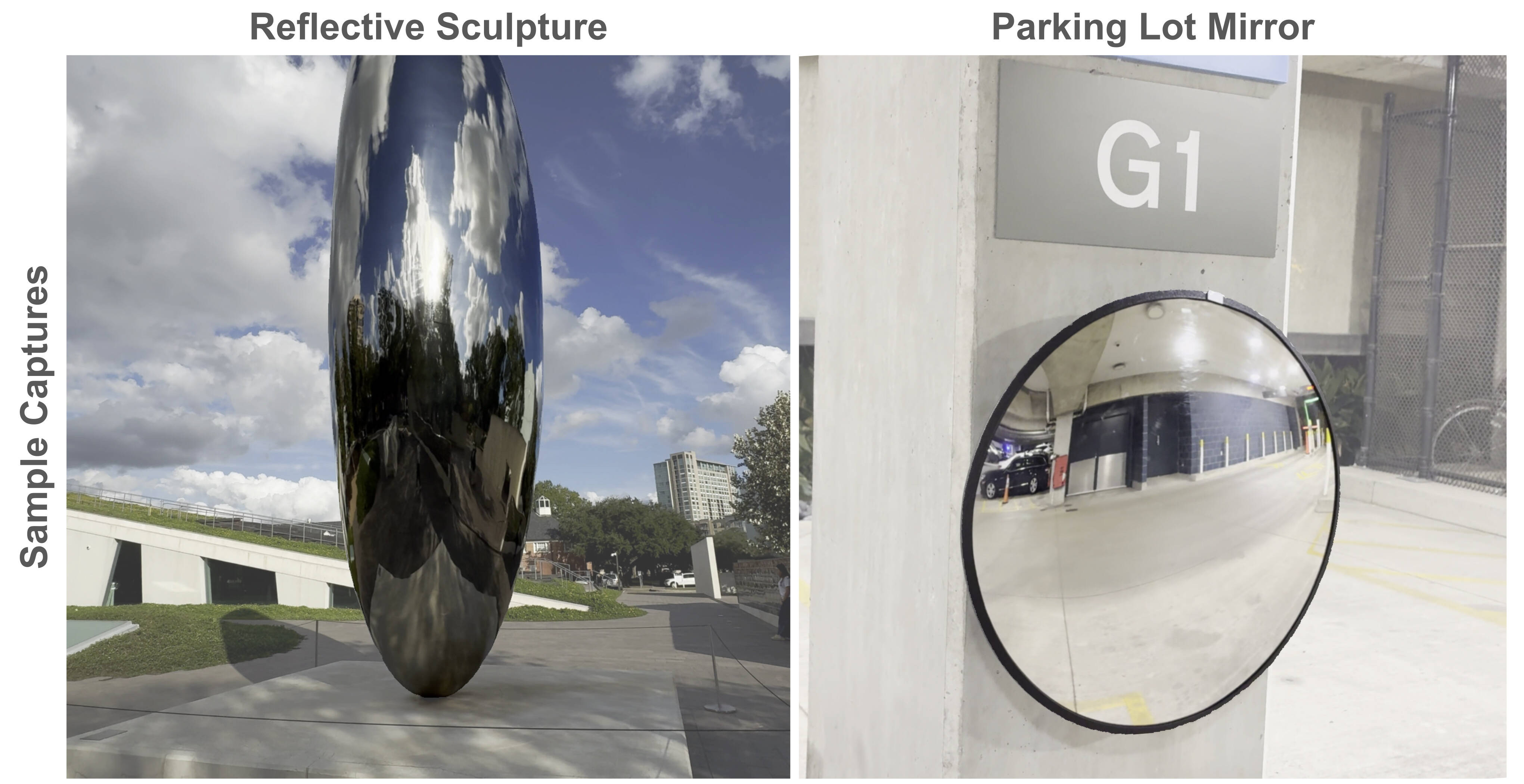}
    \caption{Sample Captures of the Shiny Real Dataset: Reflective Sculpture (Left) and Parking Lot Mirror (Right)}
    \label{fig:shiny-real-supp}
\end{figure}

\paragraph{Capture pipeline.}
Each scene is recorded as a handheld video with the operator walking in an arc around one side of the reflective object. Frames are subsampled to 12\,fps, scored by Laplacian variance, and the 350 sharpest frames are retained subject to a maximum temporal gap constraint to ensure uniform scene coverage. All frames are center-cropped and resized to $1200\!\times\!1200$ pixels before being passed to COLMAP, which is run with a \texttt{SIMPLE\_RADIAL} camera model, sequential feature matching with 30-frame overlap and quadratic extension, and standard bundle adjustment followed by image undistortion. Per-frame segmentation masks are obtained via a SAM-based pipeline to isolate the reflective surface, preventing background pixels from contaminating the reflection loss during training. Sample captures are shown in Fig.~\ref{fig:shiny-real-supp}

\paragraph{Effect of object scale on capture quality.}
The dominant factor governing real-capture quality is the angular size of the operator relative to the reflective surface. The \textbf{Reflective Sculpture} stands several meters tall, so the operator and phone subtend a negligible solid angle in the reflection. A wide handheld arc therefore yields a well-conditioned COLMAP trajectory with diverse reflection directions, and the recovered environment map faithfully captures sky structure, clouds, and surrounding building contours. The \textbf{Parking Lot Mirror}, by contrast, is small: filming from directly in front causes the operator's hands and body to dominate the reflection, and their non-rigid motion violates COLMAP's static-scene assumption, corrupting pose estimation. Capture must instead be performed at strong grazing angles from one side, producing a narrow, asymmetric trajectory with limited directional coverage. This compresses the range of observable reflection directions, reducing cubemap supervision to a narrow angular band and yielding a lower-frequency environment map---while leaving surface normals and appearance reconstruction largely unaffected. Together, these two scenes illustrate a clear practical guideline: PanoLess performs best when the operator is small relative to the reflector, enabling wide-baseline capture without self-occlusion or pose degeneracy.

\section{Partial Shiny Blender Dataset Construction}
\label{sec:partial-shiny-blender}

\begin{figure}[!th]
    \centering
    \includegraphics[width=0.6\linewidth]{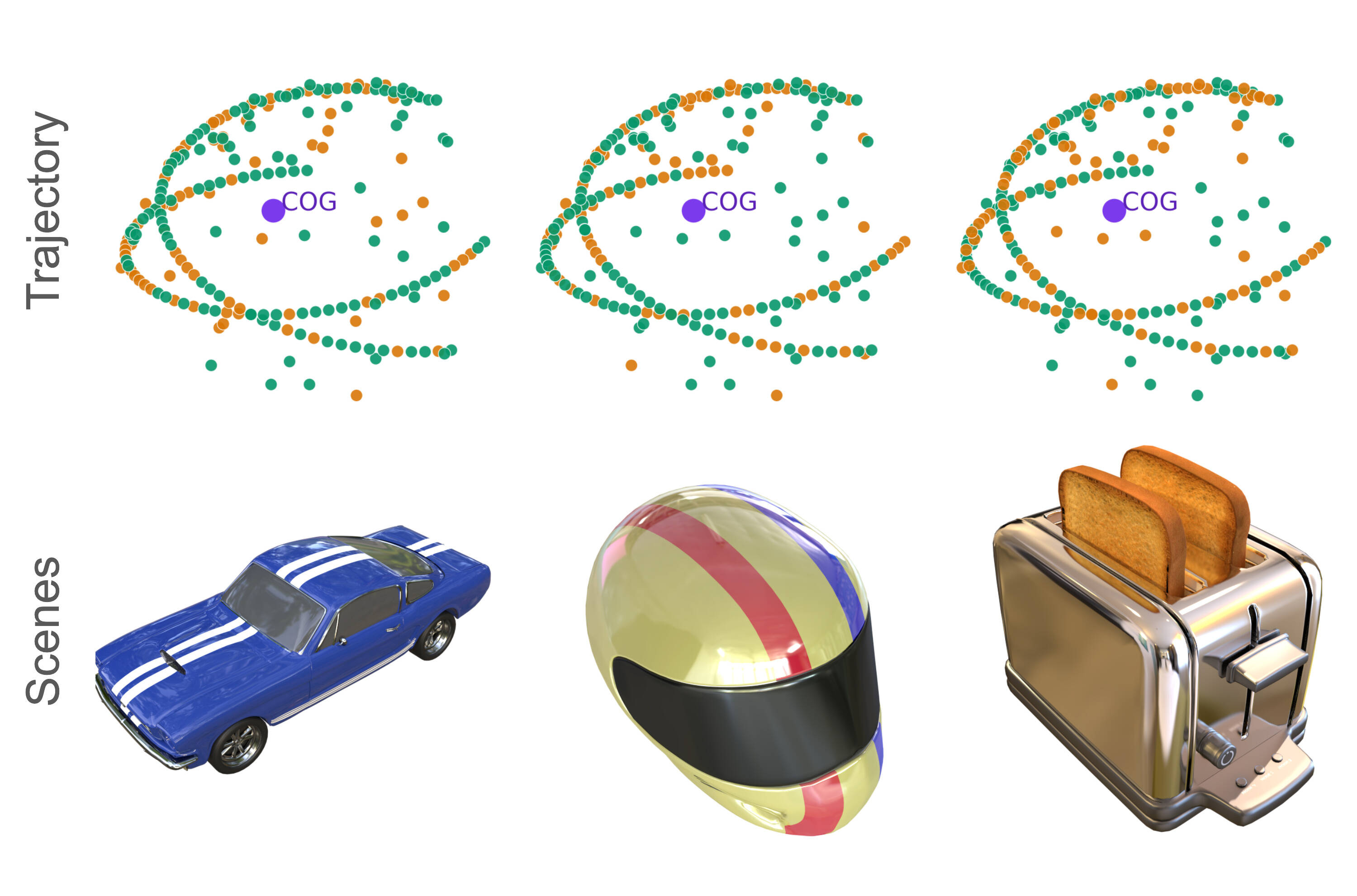}
    \caption{Filtered camera trajectories for \textbf{Partial Shiny Blender}. Training views (orange) and test views (green) are concentrated near the intersection of the original plus-shaped path, providing both horizontal and vertical angular coverage within a single frontal hemisphere. The center of Gaussians (COG) is shown in purple.}
    \label{fig:shiny-blender-traj}
\end{figure}

The Shiny Blender dataset provides dense $360^{\circ}$ camera coverage of six objects with mixed specularity, captured along a looped trajectory that crosses itself once at a central intersection point. Since full-sphere coverage falls outside our target problem setting, we construct \textbf{Partial Shiny Blender} by retaining only cameras in the vicinity of this intersection. This choice is deliberate: the crossing point is the only region of the trajectory where both horizontal and vertical viewpoint variation co-occur, so it yields richer angular diversity than a simple arc despite covering only a frontal hemisphere of the full sphere. Fig.~\ref{fig:shiny-blender-traj} shows the resulting camera distribution after filtering.

\paragraph{Train/test re-split.}
The original train/test split was designed for full-coverage training and is poorly suited to the partial-view regime: naively applying it would leave the training set severely undersampled relative to evaluation. We therefore pool all cameras that survive the hemisphere filter and re-split them randomly, assigning $40\%$ to training and $60\%$ to evaluation. We report results across five objects from the dataset, which together span a broad range of specular-to-diffuse ratios.

\section{Partial Shiny Blender Qualitative Results}
\label{sec:partial-shiny-blender-results}

\begin{figure}[!th]
    \centering
    \includegraphics[width=\linewidth]{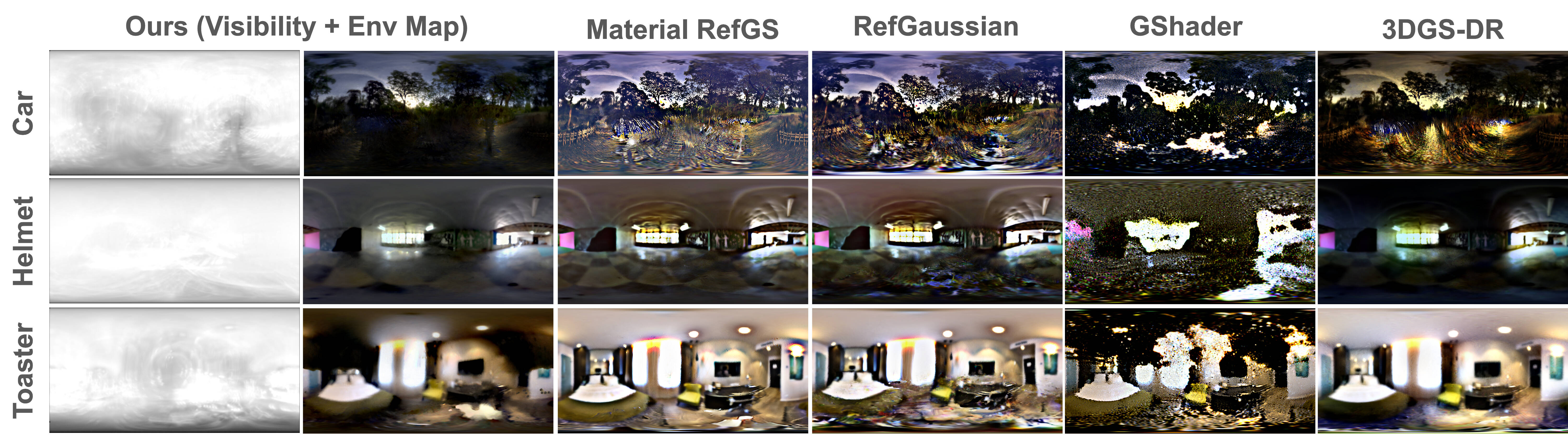}
    \caption{Recovered environment maps on \textbf{Partial Shiny Blender}. PanoLess recovers plausible coarse illumination structure across all scenes, though with lower fidelity than on the mirror-like Shiny Partial scenes. This is a direct consequence of weaker cubemap gradient signal when diffuse albedo is prominent, as discussed below.}
    \label{fig:shiny-blender-env}
\end{figure}

\begin{figure}[!th]
    \centering
    \includegraphics[width=\linewidth]{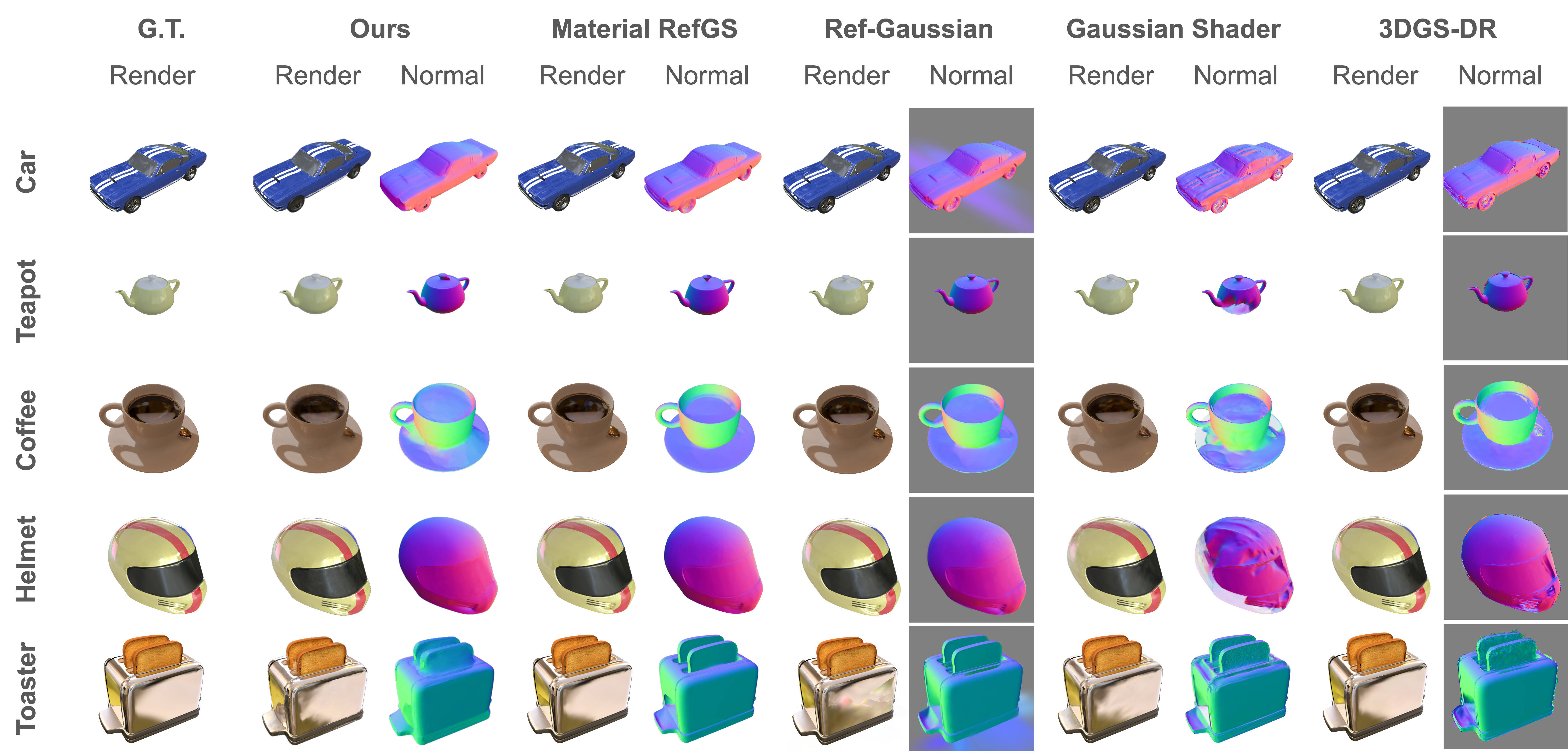}
    \caption{Novel-view renders and surface normals on \textbf{Partial Shiny Blender}. Despite the mixed-specularity materials falling outside our primary design regime, PanoLess produces geometrically accurate surface normals and competitive novel-view renders across all scenes.}
    \label{fig:shiny-blender-renders}
\end{figure}

Figs.~\ref{fig:shiny-blender-env} and~\ref{fig:shiny-blender-renders} present the recovered environment maps, surface normals, and novel-view renders for all Partial Shiny Blender scenes.

Despite operating outside its primary design regime, PanoLess recovers more structured and spatially coherent environment maps than all baselines on the majority of scenes. The \textit{Car} and \textit{Helmet} scene benefits most: specular windows and chrome trim provide sufficient high-frequency reflection signal to constrain the cubemap even under partial coverage. On more diffuse objects, however, the recovered environment maps are noticeably darker and lower-frequency than on Shiny Partial. This is a systematic consequence of the unit-reflectance assumption in our compositing model: when diffuse albedo is prominent, the Gaussian representation absorbs surface appearance into the diffuse channel, diverting photometric gradients away from the cubemap and reducing the effective supervision it receives. Since ground-truth environment maps are unavailable for this dataset, comparison remains qualitative, but the pattern is consistent and expected. PanoLess is designed for mirror-like surfaces; reduced environment map fidelity on mixed-specularity objects reflects this design choice rather than a fundamental limitation of the method.

\section{Visibility-Guided Post-Processing}
\label{sec:supp-vis}
The visibility map $\mathcal{V}$ can clean up unobserved environment directions. Thresholding $\mathcal{V}$ with Otsu's method separates observed from unsupported directions; a feathered Gaussian blur applied to the low-confidence directions smooths the noise the photometric loss leaves in unconstrained texels while leaving observed regions untouched. On \textit{Cola}, this improves every environment-map metric (Tab.~\ref{tab:vis_blur}).

\begin{table}[!th]
  \centering
  \caption{\textbf{Visibility-guided post-processing.} Otsu-thresholded visibility blur improves every envmap metric on \textit{Cola}. \colorbox{bestbg}{\textbf{Best}} highlighted.}
  \label{tab:vis_blur}
  \setlength{\tabcolsep}{6pt}
  \small
  \begin{tabular}{l cccc}
    \toprule
    \textit{Cola} envmap & PSNR\,$\uparrow$ & SSIM\,$\uparrow$ & LPIPS\,$\downarrow$ & L1\,$\downarrow$ \\
    \midrule
    Raw        & 15.32 & 0.678 & 0.469 & 0.132 \\
    + Vis-blur & \first{15.60} & \first{0.710} & \first{0.459} & \first{0.128} \\
    \bottomrule
  \end{tabular}
\end{table}

\begin{figure}[!th]
    \centering
    \includegraphics[width=0.65\linewidth]{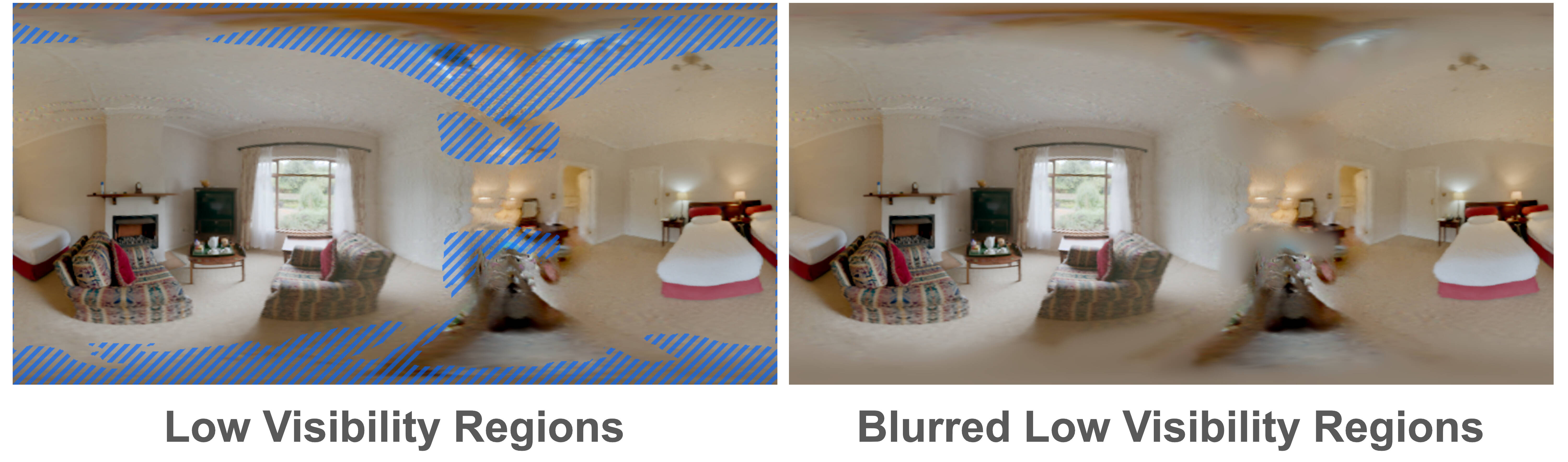}
    \caption{\textbf{Visibility identifies low-confidence regions.} Left: Otsu-thresholded visibility mask (stripes) on the recovered environment map. Right: the same map after a feathered Gaussian blur in the low-confidence directions.}
    \label{fig:inpaint}
\end{figure}

\section{Pose-Jitter Sensitivity}
\label{sec:supp-pose}
We quantify PanoLess's sensitivity to camera-pose error on \textit{Vase}. Each training-camera rotation is perturbed by a fixed random angle $\sigma$ and the model retrained, evaluating the recovered environment map against the unchanged ground truth (single seed). The fixed-$\sigma$ sweep bounds the per-camera degradation and isolates the linear dependence predicted by Eq.~(6) of the main paper. Envmap PSNR falls sharply---from $18.4$\,dB at $\sigma{=}0^\circ$ to $12.0$\,dB at $\sigma{=}0.5^\circ$---then plateaus. With no roughness to blur directional error, even sub-degree pose noise misdirects reflection queries and corrupts the recovered illumination.

\begin{table}[!th]
  \centering
  \caption{\textbf{Pose-jitter sensitivity} on \textit{Vase}. Recovered-envmap quality against the unchanged ground truth as each training camera's rotation is perturbed by a fixed angle $\sigma$ (single seed). \colorbox{bestbg}{\textbf{Best}} highlighted.}
  \label{tab:pose-jitter}
  \setlength{\tabcolsep}{6pt}
  \small
  \begin{tabular}{l cccc}
    \toprule
    $\sigma$ (deg) & PSNR\,$\uparrow$ & SSIM\,$\uparrow$ & LPIPS\,$\downarrow$ & L1\,$\downarrow$ \\
    \midrule
    0.0 & \first{18.41} & \first{0.637} & \first{0.490} & \first{0.083} \\
    0.5 & 12.04 & 0.341 & 0.599 & 0.218 \\
    1.0 & 11.71 & 0.319 & 0.616 & 0.224 \\
    \bottomrule
  \end{tabular}
\end{table}

\begin{figure}[!th]
    \centering
    \includegraphics[width=\linewidth]{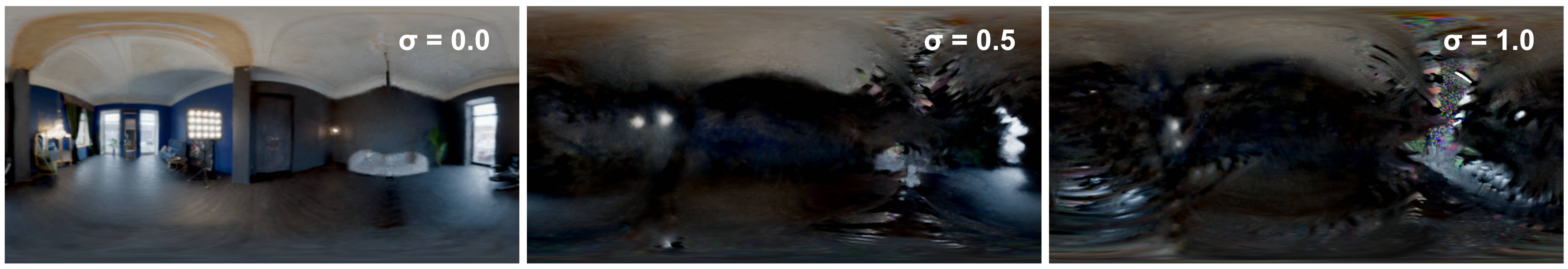}
    \caption{\textbf{Pose-jitter sensitivity on \textit{Vase}.} Recovered environment maps as training-camera rotations are perturbed by $\sigma \in \{0^\circ, 0.5^\circ, 1^\circ\}$. Even sub-degree pose error visibly degrades the recovered illumination.}
    \label{fig:pose-jitter}
\end{figure}

\end{document}